\documentstyle[11pt,times,breakcites,authdate,fleqn]{article}
\author{Philip Resnik \\
Sun Microsystems Laboratories \\
Two Elizabeth Drive \\
Chelmsford, MA 01824-4195 USA \\
{\tt philip.resnik@east.sun.com}}
\title{Disambiguating Noun Groupings with Respect to WordNet Senses}
\date{}

\newcommand{\mytabs}{............\=...............\=...............\=......
\kill}
\newcommand{\ignore}[1]{}

\newcommand{\wclass}[1]{\mbox{$\langle \mbox{{\small {\tt #1}}} \rangle$}}
\newcommand{\Prob}{{\rm Pr}}

\newcommand{\Sim}{\mbox{\rm sim}}
\newcommand{\isa}{\mbox{\sc is-a}}
\newcommand{\tinytype}{scriptsize}
\newcommand{\extraspace}{}

\begin{document}
\thispagestyle{empty}
\maketitle

\begin{center}
\begin{small}
{\bf Abstract}
\end{small}
\end{center}

\begin{small}
\begin{quote}
Word groupings useful for language processing tasks are increasingly
available, as thesauri appear on-line, and as distributional word clustering
techniques improve.  However, for many tasks, one is interested in
relationships among word {\em senses}, not words.  This paper presents a
method for automatic sense disambiguation of nouns appearing within sets of
related nouns --- the kind of data one finds in on-line thesauri, or as the
output of distributional clustering algorithms.  Disambiguation is performed
with respect to WordNet senses, which are fairly fine-grained; however, the
method also permits the assignment of higher-level WordNet categories rather
than sense labels.  The method is illustrated primarily by example, though
results of a more rigorous evaluation are also presented.
\end{quote}
\end{small}

\section{Introduction}
\label{sec:introduction}

Word groupings useful for language processing tasks are increasingly
available, as thesauri appear on-line, and as distributional techniques become
increasingly widespread (e.g.
\cite{savitch1992,brill91,brown1992,grefenstette1994,mckeown1993,pereira1993,schuetze1993:wordspace}).
However, for many tasks, one is interested in relationships among word {\em
senses}, not words.  Consider, for example, the cluster containing {\em
attorney}, {\em counsel}, {\em trial}, {\em court}, and {\em judge}, used by
Brown et al. \shortcite{brown1992} to illustrate a ``semantically sticky''
group of words.  As is often the case where sense ambiguity is involved, we as
readers impose the most coherent interpretation on the words within the group
without being aware that we are doing so. Yet a computational system has no
choice but to consider other, more awkward possibilities --- for example, this
cluster might be capturing a distributional relationship between advice (as
one sense of {\em counsel}) and royalty (as one sense of {\em court}).  This
would be a mistake for many applications, such as query expansion in
information retrieval, where a surfeit of false connections can outweigh the
benefits obtained by using lexical knowledge.

One obvious solution to this problem would be to extend distributional
grouping methods to word senses. For example, one could construct vector
representations of senses on the basis of their co-occurrence with words or
with other senses.  Unfortunately, there are few corpora annotated with word
sense information, and computing reliable statistics on word senses rather
than words will require more data, rather than less.\footnote{Actually, this
depends on the fine-grainedness of sense distinctions; clearly one could
annotate corpora with very high level semantic distinctions For example,
Basili et al. \shortcite{basili1994} take such a coarse-grained approach,
utilizing on the order of 10 to 15 semantic tags for a given domain.  I assume
throughout this paper that finer-grained distinctions than that are
necessary.} Furthermore, one widely available example of a large, manually
sense-tagged corpus --- the WordNet group's annotated subset of the Brown
corpus\footnote{Available by anonymous ftp to clarity.princeton.edu as {\tt
pub/wn1.4semcor.tar.Z}; WordNet is described by Miller et al.
\shortcite{miller1990}.} --- vividly illustrates the difficulty in obtaining
suitable data.  It is quite small, by current corpus standards (on the order
of hundreds of thousands of words, rather than millions or tens of millions);
the direct annotation methodology used to create it is labor intensive (Marcus
et al. \shortcite{marcus-et-al93} found that direct annotation takes twice as
long as automatic tagging plus correction, for part-of-speech annotation); and
the output quality reflects the difficulty of the task (inter-annotator
disagreement is on the order of 10\%, as contrasted with the approximately 3\%
error rate reported for part-of-speech annotation by Marcus et al.).

There have been some attempts to capture the behavior of semantic categories
in a distributional setting, despite the unavailability of sense-annotated
corpora.  For example, Hearst and Sch\"utze \shortcite{hearst1993} take steps
toward a distributional treatment of WordNet-based classes, using Sch\"utze's
\shortcite{schuetze1993:wordspace} approach to constructing
vector representations from a large co-occurrence matrix.  Yarowsky's
\shortcite{yarowsky1992} algorithm for sense disambiguation
can be thought of as a way of determining how Roget's thesaurus categories
behave with respect to contextual features.  And my own treatment of
selectional constraints \cite{resnik:dissertation} provides a way to describe
the plausibility of co-occurrence in terms of WordNet's semantic categories,
using co-occurrence relationships mediated by syntactic structure.  In each
case, one begins with known semantic categories (WordNet synsets, Roget's
numbered classes) and non-sense-annotated text, and proceeds to a
distributional characterization of semantic category behavior using
co-occurrence relationships.

This paper begins from a rather different starting point.  As in the
above-cited work, there is no presupposition that sense-annotated text is
available.  Here, however, I make the assumption that word groupings have been
obtained through some black box procedure, e.g. from analysis of unannotated
text, and the goal is to annotate the words within the groupings {\em post
hoc\/} using a knowledge-based catalogue of senses.  If successful, such an
approach has obvious benefits: one can use whatever sources of good word
groupings are available --- primarily unsupervised word clustering methods,
but also on-line thesauri and the like --- without folding in the complexity
of dealing with word senses at the same time.\footnote{An alternative worth
mentioning, however, is the distributional approach of Pereira et al.
\shortcite{pereira1993}: within their representational scheme, distributionally
defined word senses emerge automatically in the form of cluster centroids.}
The resulting sense groupings should be useful for a variety of purposes,
although ultimately this work is motivated by the goal of sense disambiguation
for unrestricted text using unsupervised methods.

\section{Disambiguation of Word Groups}
\label{sec:disambiguation}

\subsection{Problem statement}
\label{sec:problem}

Let us state the problem as follows.  We are given a set of words $W =
\{w_1,\ldots,w_n\}$, with each word $w_i$ having an associated set $S_i =
\{s_{i,1},\ldots,s_{i,m}\}$ of possible senses.  We assume that there exists
some set $W' \subseteq \bigcup S_i$, representing the set of word {\em
senses\/} that an ideal human judge would conclude belong to the group of
senses corresponding to the word grouping $W$.  The goal is then to define a
membership function $\varphi$ that takes $s_{i,j}$, $w_i$, and $W$ as its
arguments and computes a value in $[0,1]$, representing the confidence with
which one can state that sense $s_{i,j}$ belongs in sense grouping
$W'$.\footnote{One could also say that $\varphi$ defines a fuzzy set.} Note
that, in principle, nothing precludes the possibility that multiple senses of
a word are included in $W'$.

\extraspace
\begin{quote}
\noindent {\bf Example.}  Consider the following word group:\footnote{This
example comes from Sch\"utze's \shortcite{schuetze1993:wordspace} illustration
of how his algorithm determines nearest neighbors in a sublexical
representation space; these are the ten words representationally most similar
to {\em burglar}, based on a corpus of newspaper text.}
\begin{itemize}
\item[] burglars thief rob mugging stray robbing lookout chase crate thieves
\end{itemize}
Restricting our attention to noun senses in WordNet, only {\em lookout\/} and
{\em crate\/} are polysemous.  Treating this word group as $W$, one would
expect $\varphi$ to assign a value of 1 to the unique senses of the monosemous
words, and to assign a high value to {\em lookout}'s sense as
\begin{itemize}
\item[] lookout, lookout man, sentinel, sentry, watch, scout:
      a person employed to watch for something to happen  .
\end{itemize}
Low (or at least lower) values of $\varphi$ would be expected for the senses
of {\em lookout\/} that correspond to an observation tower, or to the activity
of watching.  {\em Crate}'s two WordNet senses correspond to the physical
object and the quantity (i.e., {\em crateful}, as in ``a crateful of
oranges''); my own intuition is that the first of these would more properly be
included in $W'$ than the second, and should therefore receive a higher value
of $\varphi$, though of course neither I nor any other individual really
constitutes an ``ideal human judge.''
\end{quote}

\subsection{Computation of Semantic Similarity}
\label{sec:similarity}

The core of the disambiguation algorithm is a computation of semantic
similarity using the WordNet taxonomy, a topic recently investigated by a
number of people \cite{leacock1994:ms,resnik1995:ijcai,sussna1993}.  In this
paper, I restrict my attention to WordNet's \isa\ taxonomy for nouns, and take
an approach in which semantic similarity is evaluated on the basis of the
information content shared by the items being compared.

The intuition behind the approach is simple: the more similar two words are,
the more informative will be the most specific concept that subsumes them
both.  (That is, their least upper bound in the taxonomy; here a {\em
concept\/} corresponds to a WordNet synset.)  The traditional method of
evaluating similarity in a semantic network by measuring the path length
between two nodes \cite{Lee93,Rada89b} also captures this, albeit indirectly,
when the semantic network is just an \isa\  hierarchy: if the minimal path of
\isa\  links between two nodes is long, that means it is necessary to go high
in
the taxonomy, to more abstract concepts, in order to find their least upper
bound.  However, there are problems with the simple path-length definition of
semantic similarity, and experiments using WordNet show that other measures of
semantic similarity, such as the one employed here, provide a better match to
human similarity judgments than simple path length does
\cite{resnik1995:ijcai}.

Given two words $w_1$ and $w_2$, their semantic similarity is
calculated as
\begin{eqnarray}
\Sim(w_1,w_2) & = &
  \max_{c \:\in\: \mbox{\rm subsumers}(w_1,w_2)} \left[- \log \Prob(c) \right],
  \label{eq:similarity}
\end{eqnarray}
where $\mbox{\rm subsumers}(w_1,w_2)$ is the set of WordNet synsets that
subsume (i.e., are ancestors of) both $w_1$ and $w_2$, in any sense of either
word.  The concept $c$ that maximizes the expression in~(\ref{eq:similarity})
will be referred to as the {\em most informative subsumer} of $w_1$ and $w_2$.
Although there are many ways to associate probabilities with taxonomic
classes, it is reasonable to require that concept probability be
non-decreasing as one moves higher in the taxonomy; i.e., that
$c_1$~\isa~$c_2$ implies $\Prob(c_2)
\geq \Prob(c_1)$.  This guarantees that ``more abstract'' does indeed mean
``less informative,'' defining informativeness in the traditional way in terms
of log likelihood.

Probability estimates are derived from a corpus by computing
\begin{eqnarray}
\mbox{\rm freq}(c) & = & \sum_{n \in \mbox{words}(c)} \mbox{\rm count}(n),
	\label{eq:estimation}
\end{eqnarray}
where $\mbox{words}(c)$ is the set of nouns having a sense subsumed by concept
$c$.  Probabilities are then computed simply as relative frequency:
\begin{eqnarray}
\hat{\Prob}(c) & = & \frac{\mbox{\rm freq}(c)}{N},
	\label{eq:rel_freq}
\end{eqnarray}
where $N$ is the total number of noun instances observed.  Singular and plural
forms are counted as the same noun, and nouns not covered by WordNet are
ignored.  Although the WordNet noun taxonomy has multiple root nodes, a
single, ``virtual'' root node is assumed to exist, with the original root
nodes as its children.  Note that by equations~(\ref{eq:similarity})
through~(\ref{eq:rel_freq}), if two senses have the virtual root node as their
only upper bound then their similarity value is 0.

\extraspace
\begin{quote}
\noindent {\bf Example.} The following table shows the semantic similarity
computed for several word pairs, in each case shown with the most informative
subsumer.\footnote{For readability, WordNet synsets are described either by
symbolic labels (such as \wclass{person}) or by long descriptions (constructed
from the words in the synset and/or the immediate parent synset and/or the
description field from the lexical database).  However, in implementations
described here, all WordNet synsets are identified by a unique numerical
identifier.} Probabilities were estimated using the Penn Treebank version of
the Brown corpus.  The pairs come from an example given by Church and Hanks
\shortcite{chur1989b}, illustrating the words that human
subjects most frequently judged as being associated with the word {\em
doctor}.  (The word {\em sick\/} also appeared on the list, but is excluded
here because it is not a noun.)
\begin{small}
\begin{center}
\extraspace
\begin{tabular}{|l|l|c|l|}\hline
Word 1	& Word 2	& Similarity	& Most Informative Subsumer \\
\hline\hline
doctor	& nurse		& 9.4823 	& \wclass{health professional}\\\hline
doctor	& lawyer	& 7.2240 	& \wclass{professional person}\\\hline
doctor	& man		& 2.9683 	& \wclass{person, individual}\\\hline
doctor	& medicine	& 1.0105 	& \wclass{entity}	\\\hline
doctor	& hospital	& 1.0105 	& \wclass{entity}	\\\hline
doctor	& health	& 0.0 		& {\em virtual root}	\\\hline
doctor	& sickness	& 0.0 		& {\em virtual root}	\\\hline
\end{tabular}
\extraspace
\end{center}
\end{small}
Doctors are minimally similar to medicine and hospitals, since these things
are all instances of ``something having concrete existence, living or
nonliving'' (WordNet class \wclass{entity}), but they are much more similar to
lawyers, since both are kinds of professional people, and even more similar to
nurses, since both are professional people working specifically within the
health professions.  Notice that similarity is a more specialized notion than
association or relatedness: doctors and sickness may be highly associated, but
one would not judge them to be particularly {\em similar\/}.
\end{quote}
\extraspace

\subsection{Disambiguation Algorithm}
\label{sec:algorithm}

The disambiguation algorithm for noun groups is inspired by the observation
that when two polysemous words are similar, their most informative subsumer
provides information about which sense of each word is the relevant one.  In
the above table, for example, both {\em doctor\/} and {\em nurse\/} are
polysemous: WordNet records {\em doctor\/} not only as a kind of health
professional, but also as someone who holds a Ph.D., and {\em nurse\/} can
mean not only a health professional but also a nanny.  When the two words are
considered together, however, the shared element of meaning for the two
relevant senses emerges in the form of the most informative subsumer.  It may
be that other pairings of possible senses also share elements of meaning (for
example, {\em doctor/Ph.D.\/} and {\em nurse/nanny\/} are both descendants of
\wclass{person, individual}).  However, in cases like those illustrated above,
the more specific or informative the shared ancestor is, the more strongly it
suggests which senses come to mind when the words are considered together.
The working hypothesis in this paper is that this holds true in general.

Turning that observation into an algorithm requires two things: a way to
assign credit to word senses based on similarity with co-occurring words, and
a tractable way to generalize to the case where more than two polysemous words
are involved.
The algorithm given in Figure~\ref{fig:algorithm}
does both quite straightforwardly.
\begin{figure}
\begin{scriptsize}
\begin{quote}
\noindent {\bf Algorithm.} Given $W = \{w[1],\ldots,w[n]\}$, a set of nouns:
\begin{tabbing}
....\=....\=.......\=......\=......\=...... \kill
\>  for i and j = 1 to n, with i < j					\\
\>  \{									\\
\>\>    v[i, j] = sim(w[i], w[j])					\\
\>\>    c[i, j] = the most informative subsumer for w[i] and w[j]	\\
\>\>									\\
\>\>    for k = 1 to num\_senses(w[i])					\\
\>\>\>      if c[i, j] is an ancestor of sense[i, k]			\\
\>\>\>\>        increment support[i, k] by v[i, j]			\\
\>\>									\\
\>\>    for k' = 1 to num\_senses(w[j])					\\
\>\>\>      if c[i, j] is an ancestor of sense[j, k']			\\
\>\>\>\>        increment support[j, k'] by v[i, j]			\\
\>\>									\\
\>\>    increment normalization[i] by v[i, j]				\\
\>\>    increment normalization[j] by v[i, j]				\\
\>  \}									\\
\>									\\
\>  for i = 1 to n							\\
\>\>    for k = 1 to num\_senses(w[i])					\\
\>\>    \{								\\
\>\>\>      if (normalization[i] > 0.0)					\\
\>\>\>\>        phi[i, k] = support[i, k] / normalization[i]		\\
\>\>\>      else							\\
\>\>\>\>        phi[i, k] = 1 / num\_senses(w[i])			\\
\>\>    \}
\end{tabbing}
\end{quote}
\end{scriptsize}
\caption{Disambiguation algorithm for noun groupings \label{fig:algorithm}}
\end{figure}

This algorithm considers the words in $W$ pairwise, avoiding the tractability
problems in considering all possible combinations of senses for the group
($O(m^n)$ if each word had $m$ senses).  For each pair considered, the most
informative subsumer is identified, and this pair is only considered as
supporting evidence for those senses that are descendants of that concept.
Notice that by equation~(\ref{eq:similarity}), {\tt support[i,k]} is a sum of
log probabilities, and therefore preferring senses with high support is
equivalent to optimizing a product of probabilities.  Thus considering words
pairwise in the algorithm reflects a probabilistic independence assumption.

\begin{quote}
\noindent {\bf Example.}  The most informative subsumer for {\em doctor\/}
and {\em nurse} is \wclass{health professional}, and therefore that pairing
contributes support to the sense of {\em doctor\/} as an M.D., but not a
Ph.D.  Similarly, it contributes support to the sense of {\em nurse\/} as a
health professional, but not a nanny.
\end{quote}

The amount of support contributed by a pairwise comparison is proportional to
how informative the most informative subsumer is.  Therefore the evidence for
the senses of a word will be influenced more by more similar words and less by
less similar words.  By the time this process is completed over all pairs,
each sense of each word in the group has had the potential of receiving
supporting evidence from a pairing with every other word in the group.  The
value assigned to that sense is then the proportion of support it {\em did\/}
receive, out of the support possible.  (The latter is kept track of by array
{\tt normalization} in the pseudocode.)

\bigskip
\noindent {\bf Discussion.} The intuition behind this algorithm is
essentially the same intuition exploited by Lesk \shortcite{lesk1986}, Sussna
\shortcite{sussna1993}, and others: the most plausible assignment of senses to
multiple co-occurring words is the one that maximizes relatedness of meaning
among the senses chosen.  Here I make an explicit comparison with Sussna's
approach, since it is the most similar of previous work.

Sussna gives as an example of the problem he is solving the following
paragraph from the corpus of 1963 {\em Time\/} magazine articles used in
information retrieval research (uppercase in the {\em Time\/} corpus,
lowercase here for readability; punctuation is as it appears in the original
corpus):
\begin{quote}
the allies after nassau in december 1960, the u.s . first proposed to help
nato develop its own nuclear strike force . but europe made no attempt to
devise a plan . last week, as they studied the nassau accord between president
kennedy and prime minister macmillan, europeans saw emerging the first
outlines of the nuclear nato that the u.s . wants and will support .  it all
sprang from the anglo-u.s .  crisis over cancellation of the bug-ridden
skybolt missile, and the u.s . offer to supply britain and france with the
proved polaris (time, dec . 28)
\end{quote}
{}From this, Sussna extracts the following noun grouping to disambiguate:
\begin{quote}
allies strike force attempt plan week accord president prime minister outlines
support crisis cancellation bug missile france polaris time
\end{quote}
These are the non-stopword nouns in the paragraph that appear in WordNet (he
used version~1.2).

The description of Sussna's algorithm for disambiguating noun groupings like
this one is similar to the one proposed here, in a number of ways: relatedness
is characterized in terms of a semantic network (specifically WordNet); the
focus is on nouns only; and evaluations of semantic similarity (or, in
Sussna's case, semantic distance) are the basis for sense selection.  However,
there are some important differences, as well.  First, unlike Sussna's
proposal, this algorithm aims to disambiguate groupings of nouns already
established (e.g. by clustering, or by manual effort) to be related, as
opposed to groupings of nouns that happen to appear near each other in running
text (which may or may not reflect relatedness based on meaning).  This
provides some justification for restricting attention to {\em similarity\/}
(reflected by the scaffolding of \isa\ links in the taxonomy), as opposed to
the more general notion of {\em association}.  Second, this difference is
reflected algorithmically by the fact that Sussna uses not only
\isa\  links but also other WordNet links such as \mbox{\sc part-of}.  Third,
unlike Sussna's algorithm, the semantic similarity/distance computation here
is not based on path length, but on information content, a choice that I have
argued for elsewhere \cite{resnik:dissertation,resnik1995:ijcai}.  Fourth, the
combinatorics are handled differently: Sussna explores analyzing all sense
combinations (and living with the exponential complexity), as well as the
alternative of sequentially ``freezing'' a single sense for each of
$w_1,\ldots,w_{i-1}$ and using those choices, assumed to be correct, as the
basis for disambiguating $w_i$.  The algorithm presented here falls between
those two alternatives.

A final, important difference between this algorithm and previous algorithms
for sense disambiguation is that it offers the possibility of assigning
higher-level WordNet categories rather than lowest-level sense labels.  It is
a simple modification to the algorithm to assign values of $\varphi$ not only
to synsets directly containing words in $W$, but to any ancestors of those
synsets --- one need only let the list of synsets associated with each word
$w_i$ (i.e., $S_i$ in the problem statement of Section~\ref{sec:problem}) also
include any synset that is an ancestor of any synset containing word $w_i$.
Assuming that \verb|num_senses(w[i])| and
\verb|sense[i,k]| are reinterpreted accordingly, the algorithm will compute
$\varphi$ not only for the synsets directly including words in $W$, but also
for any higher-level abstractions of them.

\extraspace
\begin{quote}
\noindent {\bf Example.}  Consider the word group {\em doctor}, {\em nurse},
{\em lawyer}.  If one were to include all subsuming concepts for each word,
rather than just the synsets of which they are directly members, the concepts
with non-zero values of $\varphi$ would be as follows:
\begin{itemize}
\item For {\em doctor}:
\begin{\tinytype}
\begin{tabbing}
\mytabs
 \>     1.00 \>    doctor, doc, physician, MD, Dr.: subconcept of medical
practitioner\\
 \>     1.00 \>    medical practitioner: someone who practices medicine  \\
 \>     1.00 \>    health professional: subconcept of professional\\
 \>     0.43 \>    professional: a person engaged in one of the learned
professions  \\
\end{tabbing}
\end{\tinytype}
\item For {\em nurse}:
\begin{\tinytype}
\begin{tabbing}
\mytabs
 \>     1.00 \>   nurse: one skilled in caring for the sick \\
 \>     1.00 \>   health professional: subconcept of professional\\
 \>     0.43 \>   professional: a person engaged in one of the learned
professions
\end{tabbing}
\end{\tinytype}
\item For {\em lawyer}:
\begin{\tinytype}
\begin{tabbing}
\mytabs
 \>     1.00 \>   lawyer, attorney: a professional person authorized to
practice law\\
 \>     1.00 \>   professional: a person engaged in one of the learned
professions  \\
\end{tabbing}
\end{\tinytype}
\end{itemize}
\end{quote}
\extraspace
Given assignments of $\varphi$ at all levels of abstraction, one obvious
method of semantic annotation is to assign the {\em highest-level\/} concept
for which $\varphi$ is at least as large as the sense-specific value of
$\varphi$.  For instance, in the previous example, one would assign the
annotation \wclass{health professional} to both {\em doctor\/} and {\em
nurse\/} (thus explicitly capturing a generalization about their presence in
the word group, at the appropriate level of abstraction), and the annotation
\wclass{professional} to {\em lawyer}.

\section{Examples}
\label{sec:examples}

In this section I present a number of examples for evaluation by inspection.
In each case, I give the source of the noun grouping, the grouping itself, and
for each word a description of word senses together with their values of
$\varphi$.

\extraspace
\subsection{Distributionally derived groupings}

\noindent{\bf Distributional cluster} \cite{brown1992}:
head, body, hands, eye, voice, arm, seat, hair, mouth \\
\begin{\tinytype}
\begin{tabbing}
\mytabs
Word 'head'  (17 alternatives)\\
 \>  0.0000 \>   crown, peak, summit, head, top: subconcept of upper~bound \\
 \>  0.0000 \>   principal, school~principal, head~teacher, head: educator who
has
executive authority \\
 \>  0.0000 \>   head, chief, top~dog: subconcept of leader \\
 \>  0.0000 \>   head: a user of (usually soft) drugs  \\
 \>  0.1983 \>   head: "the head of the page"; "the head of the list"  \\
 \>  0.1983 \>   beginning, head, origin, root, source: the point or place
where something begins  \\
 \>  0.0000 \>   pass, head, straits: a difficult juncture; "a pretty pass"  \\
 \>  0.0000 \>   headway, head: subconcept of progress, progression, advance \\
 \>  0.0903 \>   point, head: a V-shaped mark at one end of an arrow pointer
\\
 \>  0.0000 \>   heading, head: a line of text serving to indicate what the
passage below it is about  \\
 \>  0.0000 \>   mind, head, intellect, psyche: that which is responsible for
your thoughts and feelings\\
 \>  0.5428 \>   head: the upper or front part of the body that contains the
face and brains  \\
 \>  0.0000 \>   toilet, lavatory, can, head, facility, john, privy, bathroom\\
 \>  0.0000 \>   head: the striking part of a tool; "hammerhead"  \\
 \>  0.1685 \>   head: a part that projects out from the rest; "the head of the
nail", "pinhead"  \\
 \>  0.0000 \>   drumhead, head: stretched taut  \\
 \>  0.0000 \>   oral~sex, head: oral-genital stimulation  \\
\\
Word 'body'  (8 alternatives)\\
 \>  0.0000 \>   body: an individual 3-dimensional object that has mass \\
 \>  0.0000 \>   gathering, assemblage, assembly, body, confluence: group of
people together in one place  \\
 \>  0.0000 \>   body: people associated by some common tie or occupation  \\
 \>  0.0000 \>   body: the central message of a communication  \\
 \>  0.9178 \>   torso, trunk, body: subconcept of body~part, member \\
 \>  0.0000 \>   body, organic~structure: the entire physical structure of an
animal or human being  \\
 \>  0.0822 \>   consistency, consistence, body: subconcept of property \\
 \>  0.0000 \>   fuselage, body: the central portion of an airplane \\
\\
Word 'hands'  (10 alternatives)\\
 \>  0.0000 \>   hand: subconcept of linear~unit \\
 \>  0.0653 \>   hired~hand, hand, hired~man: a hired laborer on a farm or
ranch  \\
 \>  0.0653 \>   bridge~player, hand: "we need a 4th hand for bridge"  \\
 \>  0.0000 \>   hand, deal: the cards held in a card game by a given player at
any given time  \\
 \>  0.0000 \>   hand: a round of applause to signify approval; "give the
little lady a great big hand"  \\
 \>  0.0000 \>   handwriting, cursive, hand, script: something written by hand
\\
 \>  0.2151 \>   hand: ability; "he wanted to try his hand at singing"  \\
 \>  0.7196 \>   hand, manus, hook, mauler, mitt, paw: the distal extremity of
the superior limb  \\
 \>  0.0000 \>   hand: subconcept of pointer \\
 \>  0.0000 \>   hand: physical assistance; "give me a hand with the chores"
\\
\\
Word 'eye'  (4 alternatives)\\
 \>  0.1479 \>   center, centre, middle, heart, eye: approximately central
within some region  \\
 \>  0.1547 \>   eye: good judgment; "she has an eye for fresh talent"  \\
 \>  0.6432 \>   eye, eyeball, oculus, optic, peeper, organ~of~sight\\
 \>  0.0542 \>   eye: a small hole  or loop (as in a needle)  \\
\\
Word 'voice'    (7 alternatives)\\\\
 \>  0.0000 \>   voice: the relation of the subject of a verb to the action
that the verb denotes  \\
 \>  0.1414 \>   spokesperson, spokesman, interpreter, representative,
mouthpiece, voice\\
 \>  0.1122 \>   voice, vocalization: the sound made by the vibration of vocal
folds \\
 \>  0.2029 \>   articulation, voice: expressing in coherent verbal form; "I
gave voice to my feelings"  \\
 \>  0.3895 \>   part, voice: the melody carried by a particular voice or
instrument in polyphonic music  \\
 \>  0.0000 \>   voice: the ability to speak; "he lost his voice"  \\
 \>  0.1539 \>   voice: the distinctive sound of a person's speech; "I
recognized her voice"  \\
\\
Word 'arm'  (6 alternatives)\\
 \>  0.0000 \>   branch, subdivision, arm: an administrative division: "a
branch of Congress"  \\
 \>  0.6131 \>   arm: commonly used to refer to the whole superior limb  \\
 \>  0.0346 \>   weapon, arm, weapon~system: used in fighting or hunting  \\
 \>  0.2265 \>   sleeve, arm: attached at armhole  \\
 \>  0.1950 \>   arm: any projection that is thought to resemble an arm; "the
arm of the record player"\\
 \>  0.0346 \>   arm: the part of an armchair that supports the elbow and
forearm of a seated person  \\
\\
Word 'seat'  (6 alternatives)\\
 \>  0.0000 \>   seat: a city from which authority is exercised  \\
 \>  0.0000 \>   seat, place: a space reserved for sitting\\
 \>  0.7369 \>   buttocks, arse, butt, backside, bum, buns, can, ...\\
 \>  0.2631 \>   seat: covers the buttocks  \\
 \>  0.0402 \>   seat: designed for sitting on  \\
 \>  0.0402 \>   seat: where one sits  \\
\\
Word 'hair'  (5 alternatives)\\
 \>  0.0323 \>   hair, pilus: threadlike keratinous filaments growing from the
skin of mammals  \\
 \>  0.2313 \>   hair, tomentum: filamentous hairlike growth on a plant  \\
 \>  1.0000 \>   hair, follicular~growth: subconcept of external~body~part \\
 \>  1.0000 \>   hair, mane, head~of~hair: hair on the head  \\
 \>  1.0000 \>   hair: hairy covering of an animal or body part  \\
\\
Word 'mouth'  (5 alternatives)\\
 \>  0.0000 \>   mouth: the point where a stream issues into a larger body of
water  \\
 \>  0.0000 \>   mouth: an opening that resembles a mouth (as of a cave or a
gorge)\\
 \>  0.0613 \>   sass, sassing, backtalk, lip, mouth: an impudent or insolent
rejoinder  \\
 \>  0.9387 \>   mouth, oral~cavity: subconcept of cavity, body~cavity,
bodily~cavity \\
 \>  0.9387 \>   mouth, trap, hole, maw, yap, muzzle, snout: list includes
informal terms for "mouth"  \\

\end{tabbing}
\end{\tinytype}
This group was among classes hand-selected by Brown et al. as ``particularly
interesting.''

\vspace{0.5cm}
\ \hrulefill\
\vspace{0.5cm}

\noindent{\bf Distributional cluster} \cite{brown1992}: tie, jacket, suit
\begin{\tinytype}
\begin{tabbing}
\mytabs
Word 'tie'  (7 alternatives)\\
 \>    0.0000 \>   draw, standoff, tie, stalemate\\
 \>    0.0000 \>   affiliation, association, tie, tie-up: a social or business
relationship\\
 \>    0.0000 \>   tie, crosstie, sleeper: subconcept of brace, bracing \\
 \>    1.0000 \>   necktie, tie\\
 \>    0.0000 \>   link, linkup, tie, tie-in: something that serves to join or
link  \\
 \>    0.0000 \>   drawstring, string, tie: cord used as a fastener  \\
 \>    0.0000 \>   tie, tie~beam: used to prevent two rafters, e.g., from
spreading apart\\
\\
Word 'jacket'  (4 alternatives)\\
 \>    0.0000 \>   book~jacket, dust~cover: subconcept of promotional~material
\\
 \>    0.0000 \>   jacket~crown, jacket: artificial crown fitted over a broken
or decayed tooth  \\
 \>    0.0000 \>   jacket: subconcept of wrapping, wrap, wrapper \\
 \>    1.0000 \>   jacket: a short coat  \\
\\
Word 'suit'  (4 alternatives)\\
 \>    0.0000 \>   suit, suing: subconcept of entreaty, prayer, appeal \\
 \>    1.0000 \>   suit, suit~of~clothes: subconcept of garment \\
 \>    0.0000 \>   suit: any of four sets of 13" cards in a pack  \\
 \>    0.0000 \>   legal~action, action, case, lawsuit, suit: a judicial
proceeding\\
\end{tabbing}
\end{\tinytype}
This cluster was derived by Brown et al. using a modification of their
algorithm, designed to uncover ``semantically sticky'' clusters.

\vspace{0.5cm}
\ \hrulefill\
\vspace{0.5cm}

\noindent{\bf Distributional cluster} \cite{brown1992}:
cost, expense, risk, profitability, deferral, earmarks, capstone,
cardinality, mintage, reseller \\
\begin{\tinytype}
\begin{tabbing}
\mytabs
Word 'cost'  (2 alternatives)\\
 \>  0.5426 \>   cost, price, terms, damage: the amount of money paid for
something  \\
 \>  0.4574 \>   monetary~value, price, cost: the amount of money it would
bring if sold  \\
\\
Word 'expense'  (2 alternatives)\\
 \>  1.0000 \>   expense, expenditure, outlay, outgo, spending, disbursal,
disbursement\\
 \>  0.0000 \>   expense: a detriment or sacrifice; "at the expense of"  \\
\\
Word 'risk'  (2 alternatives)\\
 \>  0.6267 \>   hazard, jeopardy, peril, risk: subconcept of danger \\
 \>  0.3733 \>   risk, peril, danger: subconcept of venture \\
\\
Word 'profitability'  (1 alternatives)\\
 \>  1.0000 \>   profitableness, profitability: subconcept of advantage,
benefit, usefulness \\
\\
Word 'deferral'  (3 alternatives)\\
 \>  0.6267 \>   abeyance, deferral, recess: subconcept of inaction,
inactivity, inactiveness \\
 \>  0.3733 \>   postponement, deferment, deferral, moratorium: an agreed
suspension of activity  \\
 \>  0.3733 \>   deferral: subconcept of pause, wait \\
\\
Word 'earmarks'  (2 alternatives)\\
 \>  0.2898 \>   earmark: identification mark on the ear of a domestic animal
\\
 \>  0.7102 \>   hallmark, trademark, earmark: a distinguishing characteristic
or attribute  \\
\\
Word 'capstone'  (1 alternatives)\\
 \>  1.0000 \>   capstone, coping~stone, stretcher: used at top of wall  \\
\\
Word 'cardinality' \\
 \>      Not in  WordNet\\
\\
Word 'mintage'  (1 alternatives)\\
 \>  1.0000 \>   coinage, mintage, specie, metal~money: subconcept of cash \\
\\
Word 'reseller' \\
 \>      Not in  WordNet\\
\end{tabbing}
\end{\tinytype}
This cluster was one presented by Brown et al. as a randomly-selected class,
rather than one hand-picked for its coherence.  (I hand-selected it from that
group for presentation here, however.)

\vspace{0.5cm}
\ \hrulefill\
\vspace{0.5cm}

\noindent{\bf Distributional neighborhood} \cite{schuetze1993:wordspace}:
burglars, thief, rob, mugging, stray, robbing, lookout, chase, crate \\
\begin{\tinytype}
\begin{tabbing}
\mytabs
Word 'burglars'  (1 alternatives)\\
 \>    1.0000 \>   burglar: subconcept of thief, robber \\
\\
Word 'thief'  (1 alternatives)\\
 \>    1.0000 \>   thief, robber: subconcept of criminal, felon, crook, outlaw
\\
\\
Word 'mugging'  (1 alternatives)\\
 \>    1.0000 \>   battering, beating, mugging, whipping: subconcept of fight,
fighting \\
\\
Word 'stray'  (1 alternatives)\\
 \>    1.0000 \>   alley~cat, stray: homeless cat  \\
\\
Word 'lookout'  (4 alternatives)\\
 \>    0.6463 \>   lookout, lookout~man, sentinel, sentry, watch, scout\\
 \>    0.0000 \>   lookout, observation~post: an elevated post affording a wide
view  \\
 \>    0.1269 \>   lookout, observation~tower, lookout~station, observatory: \\
 \>    0.2268 \>   lookout, outlook: subconcept of look, looking~at \\
\\
Word 'chase'  (1 alternatives)\\
 \>    1.0000 \>   pursuit, chase, follow, following: the act of pursuing  \\
\\
Word 'crate'  (2 alternatives)\\
 \>    0.0000 \>   crate, crateful: subconcept of containerful \\
 \>    1.0000 \>   crate: a rugged box (usually made of wood); used for
shipping  \\
\end{tabbing}
\end{\tinytype}
As noted in Section~\ref{sec:problem}, this group represents a set of words
similar to {\em burglar}, according to Sch\"utze's method for deriving vector
representation from corpus behavior.  In this case, words {\em rob\/} and {\em
robbing\/} were excluded because they were not nouns in WordNet.  The word
{\em stray\/} probably should be excluded also, since it most likely appears
on this list as an adjective (as in ``stray bullet'').

\vspace{0.5cm}
\ \hrulefill\
\vspace{0.5cm}

\noindent{\bf Machine-generated thesaurus entry} \cite{grefenstette1994}:
method, test, mean, procedure, technique \\
\begin{\tinytype}
\begin{tabbing}
\mytabs
Word 'method'  (2 alternatives)\\
 \>    1.0000 \>   method: a way of doing something, esp. a systematic one\\
 \>    0.0000 \>   wise, method: a way of doing or being: "in no wise"; "in
this wise"\\
\\
Word 'test'  (7 alternatives)\\
 \>    0.6817 \>   trial, test, tryout: trying something to find out about it;
"ten days free trial"  \\
 \>    0.6817 \>   assay, check, test: subconcept of appraisal, assessment \\
 \>    0.0000 \>   examination, exam, test: a set of questions or exercises
evaluating skill or knowledge  \\
 \>    0.3183 \>   test, mental~test, mental~testing, psychometric~test\\
 \>    0.0000 \>   test: a hard outer covering as of some amoebas and sea
urchins  \\
 \>    0.3183 \>   test, trial: the act of undergoing testing; "he survived the
great test of battle"\\
 \>    0.3183 \>   test, trial, run: the act of testing something\\
\\
Word 'mean'  (1 alternatives)\\
 \>    1.0000 \>   mean: an average of n numbers computed by...\\
\\
Word 'procedure'  (4 alternatives)\\
 \>    1.0000 \>   procedure, process: a particular course of action intended
to achieve a results\\
 \>    1.0000 \>   operation, procedure: a process or series of acts ...
involved in a particular form of work\\
 \>    0.0000 \>   routine, subroutine, subprogram, procedure, function\\
 \>    0.0000 \>   procedure: a mode of conducting legal and parliamentary
proceedings\\
\\
Word 'technique'  (2 alternatives)\\
 \>    1.0000 \>   technique: a technical method  \\
 \>    0.0000 \>   proficiency, facility, technique: skillfulness deriving from
practice and familiarity\\
\end{tabbing}
\end{\tinytype}
I chose this grouping at random from a thesaurus created automatically by
Grefenstette's syntactico-distributional methods, using the MED corpus of
medical abstracts as its source.  The group comes from from the thesaurus
entry for the word {\em method}.  Note that {\em mean\/} probably should be
{\em means}.

\extraspace
\subsection{Thesaurus Classes}

There is a tradition in sense disambiguation of taking particularly ambiguous
words and evaluating a system's performance on those words.  Here I look at
one such case, the word {\em line}; the goal is to see what sense the
algorithm chooses when considering the word in the contexts of each of the
Roget's Thesaurus classes in which it appears, where a ``class'' includes all
the nouns in one of the numbered categories.\footnote{I am grateful to Mark
Lauer for his kind assistance with the thesaurus.} The following list
provides brief descriptions of the 25 senses of {\em line\/} in WordNet:

\extraspace
\begin{scriptsize}
\begin{enumerate}
\item wrinkle, furrow, crease, crinkle, seam, line: "His face has many
wrinkles"
\item line: a length (straight or curved) without breadth or thickness
\item line, dividing~line: "there is a narrow line between sanity and insanity"
\item agate~line, line: space for one line of print used to measure advertising
\item credit~line, line~of~credit, line: the maximum credit that a  customer is
allowed
\item line: in games or sports; a mark indicating positions or bounds of the
playing area
\item line: a spatial location defined by a real or imaginary unidimensional
extent
\item course, line: a connected series of events or actions or developments
\item line: a formation of people or things one after (or beside) another
\item lineage, line, line~of~descent, descent, bloodline, blood~line, blood,
pedigree
\item tune, melody, air, strain, melodic~line, line, melodic~phrase: a
succession of notes
\item line: a linear string of words expressing some idea
\item line: a mark that is long relative to its width; "He drew a line on the
chart"
\item note, short~letter, line: "drop me a line when you get there"
\item argumentation, logical~argument, line~of~thought, line~of~reasoning, line
\item telephone~line, phone~line, line: a telephone connection
\item production~line, assembly~line, line: a factory system
\item pipeline, line: a long pipe used to transport liquids or gases
\item line: a commercial organization serving as a common carrier
\item line, railway~line, rail~line: railroad track and roadbed
\item line: something long and thin and flexible
\item cable, line, transmission~line: electrical conductor connecting
telephones or television
\item line, product~line, line~of~products, line~of~merchandise, business~line,
line~of~business
\item line: acting in conformity; "in line with" or "he got out of line" or
"toe the line"
\item occupation, business, line~of~work, line: the principal activity in your
life
\end{enumerate}
\end{scriptsize}

\extraspace
Since {\em line\/} appears in 13 of the numbered categories in Roget's
thesaurus, a full description of the values of $\varphi$ would be too large
for the present paper.  Indeed, showing all the nouns in the numbered
categories would take up too much space: they average about 70 nouns apiece.
Instead, I identify the numbered category, and give the three WordNet senses
of {\em line\/} for which $\varphi$ was greatest.

\begin{\tinytype}
\begin{tabbing}
\mytabs
{\scriptsize {\bf [\#45.] [Connecting medium.] Connection. }} \\
 \> 0.4280 \>   cable, line, transmission line\\
 \> 0.2966 \>   telephone line, phone line, line: a telephone connection  \\
 \> 0.2838 \>   line: something long and thin and flexible  \\
\\
{\scriptsize {\bf [\#69.] [Uninterrupted sequence.] Continuity. }} \\
 \> 0.3027 \>   lineage, line, line of descent\\
 \> 0.2172 \>   line: a formation of people or things one after (or beside)
another\\
 \> 0.1953 \>   course, line: a connected series of events or actions or
developments\\
\\
{\scriptsize {\bf [\#166.] Paternity.}} \\
 \> 0.5417 \>   lineage, line, line of descent, descent, bloodline, blood line,
blood, pedigree\\
 \> 0.2292 \>   pipeline, line: a long pipe used to transport liquids or
gases\\
 \> 0.1559 \>   line, product line, line of products, line of merchandise\\
\\
{\scriptsize {\bf [\#167.] Posterity.}} \\
 \> 0.3633 \>   lineage, line, line of descent, descent, bloodline, blood line,
blood, pedigree\\
 \> 0.2904 \>   line, product line, line of products, line of merchandise\\
 \> 0.2464 \>   cable, line, transmission line: electrical conductor connecting
telephones or television\\
\\
{\scriptsize {\bf [\#200.] Length.}} \\
 \> 0.5541 \>   agate line, line: space for one line of print used to measure
advertising  \\
 \> 0.0906 \>   cable, line, transmission line: electrical conductor connecting
telephones or television\\
 \> 0.0894 \>   telephone line, phone line, line: a telephone connection  \\
\\
{\scriptsize {\bf [\#203.] Narrowness. Thinness.}} \\
 \> 0.2496 \>   pipeline, line: a long pipe used to transport liquids or
gases\\
 \> 0.2141 \>   line: a linear string of words expressing some idea  \\
 \> 0.2141 \>   note, short letter, line: "drop me a line when you get there"
\\
\\
{\scriptsize {\bf [\#205.] Filament.}} \\
 \> 0.5724 \>   line: something long and thin and flexible  \\
 \> 0.1805 \>   cable, line, transmission line: electrical conductor connecting
telephones or television\\
 \> 0.1425 \>   line: in games or sports; a mark indicating positions or bounds
of the playing area  \\
\\
{\scriptsize {\bf [\#278.] Direction. }} \\
 \> 0.2083 \>   line: a spatial location defined by a real or imaginary
unidimensional extent  \\
 \> 0.1089 \>   wrinkle, furrow, crease, crinkle, seam, line: "His face has
many wrinkles"  \\
 \> 0.1031 \>   line: a length without breadth or thickness; the trace of a
moving point  \\
\\
{\scriptsize {\bf [\#413.] Melody. Concord. }} \\
 \> 0.3474 \>   note, short letter, line: "drop me a line when you get there"
\\
 \> 0.1337 \>   agate line, line: space for one line of print used to measure
advertising  \\
 \> 0.1030 \>   tune, melody, air, strain, melodic line, line, melodic phrase\\
\\
{\scriptsize {\bf [\#466.] Measurement. }} \\
 \> 0.5423 \>   cable, line, transmission line: electrical conductor connecting
telephones or television\\
 \> 0.1110 \>   argumentation, logical argument, line of thought, line of
reasoning, line\\
 \> 0.0969 \>   agate line, line: space for one line of print used to measure
advertising  \\
\\
{\scriptsize {\bf [\#590.] Writing. }} \\
 \> 0.4743 \>   note, short letter, line: "drop me a line when you get there"
\\
 \> 0.1734 \>   cable, line, transmission line: electrical conductor connecting
telephones or television\\
 \> 0.1648 \>   tune, melody, air, strain, melodic line, line, melodic phrase\\
\\
{\scriptsize {\bf [\#597.] Poetry. }} \\
 \> 0.3717 \>   note, short letter, line: "drop me a line when you get there"
\\
 \> 0.2689 \>   tune, melody, air, strain, melodic line, line, melodic phrase\\
 \> 0.2272 \>   line: a linear string of words expressing some idea  \\
\\
{\scriptsize {\bf [\#625.] Business. }} \\
 \> 0.4684 \>   occupation, business, line of work, line: the principal
activity in your life\\
 \> 0.1043 \>   line: a commercial organization serving as a common carrier  \\
 \> 0.0790 \>   tune, melody, air, strain, melodic line, line, melodic phrase\\
\end{tabbing}
\end{\tinytype}

\extraspace
\noindent Qualitatively, the algorithm does a good job in most of
the categories.  The reader might find it an interesting exercise to try to
decide which of the 25 senses he or she would choose, especially in the cases
where the algorithm did less well (e.g. categories \#200, \#203, \#466).

\section{Formal Evaluation}
\label{sec:evaluation}

The previous section provided illustrative examples, demonstrating the
performance of the algorithm on some interesting cases.  In this section, I
present experimental results using a more rigorous evaluation methodology.

Input for this evaluation came from the numbered categories of Roget's.  Test
instances consisted of a noun group (i.e., all the nouns in a numbered
category) together with a single word in that group to be disambiguated.  To
use an example from the previous section, category \#590 (``Writing'')
contains the following:

\begin{footnotesize}
\begin{quote}
writing, chirography, penmanship, quill~driving, typewriting, writing,
manuscript, MS, these~presents, stroke~of~the~pen, dash~of~the~pen,
coupe~de~plume, line, headline, pen~and~ink, letter, uncial~writing,
cuneiform~character, arrowhead, Ogham, Runes, hieroglyphic, contraction,
Devanagari, Nagari, script, shorthand, stenography, secret~writing,
writing~in~cipher, cryptography, stenography, copy, transcript, rescript,
rough~copy, fair~copy, handwriting, signature, sign~manual, autograph,
monograph, holograph, hand, fist, calligraphy, good~hand, running~hand,
flowing~hand, cursive~hand, legible~hand, bold~hand, bad~hand, cramped~hand,
crabbed~hand, illegible~hand, scribble, ill-formed~letters,
pothooks~and~hangers, stationery, pen, quill, goose~quill, pencil, style,
paper, foolscap, parchment, vellum, papyrus, tablet, slate, marble, pillar,
table, blackboard, ink~bottle, ink~horn, ink~pot, ink~stand, ink~well,
typewriter, transcription, inscription, superscription, graphology,
composition, authorship, writer, scribe, amanuensis, scrivener, secretary,
clerk, penman, copyist, transcriber, quill~driver, stenographer, typewriter,
typist, writer~for~the~press
\end{quote}
\end{footnotesize}

\noindent Any word or phrase in that group that appears in the noun taxonomy
for WordNet would be a candidate as a test instance --- for example, {\em
line}, or {\em secret~writing}.

The test set, chosen at random, contained 125 test cases.  (Note that because
of the random choice, there were some cases where more than one test instance
came from the same numbered category.)  Two human judges were independently
given the test cases to disambiguate.  For each case, they were given the full
set of nouns in the numbered category (as shown above) together with
descriptions of the WordNet senses for the word to be disambiguated (as, for
example, the list of 25 senses for {\em line\/} given in the previous section,
though thankfully few words have that many senses!).  It was a forced-choice
task; that is, the judge was required to choose exactly one sense.  In
addition, for each judgment, the judge was required to provide a confidence
value for this decision, ranging from 0 (not at all confident) to 4 (highly
confident).

Results are presented here individually by judge.  For purposes of evaluation,
test instances for which the judge had low confidence (i.e. confidence ratings
of~0 or~1) were excluded.

For Judge~1, there were 99 test instances with sufficiently high confidence to
be considered.  As a baseline, ten runs were done selecting senses by random
choice, with the average percent correct being 34.8\%, standard deviation
3.58.  As an upper bound, Judge~2 was correct on 65.7\% of those test
instances.  The disambiguation algorithm shows considerable progress toward
this upper bound, with 58.6\% correct.

For Judge~2, there were 86 test instances with sufficiently high confidence to
be considered.  As a baseline, ten runs were done selecting senses by random
choice, with the average percent correct being 33.3\%, standard deviation
3.83.  As an upper bound, Judge~1 was correct on 68.6\% of those test
instances. Again, the disambiguation algorithm performs well, with 60.5\%
correct.

\section{Conclusions and Future Work}
\label{sec:future}

The results of the evaluation are extremely encouraging, especially
considering that disambiguating word senses to the level of fine-grainedness
found in WordNet is quite a bit more difficult than disambiguation to the
level of homographs \cite{hearst1991,cowie1992:coling}.  A note worth adding:
it is not clear that the ``exact match'' criterion --- that is, evaluating
algorithms by the percentage of exact matches of sense selection against a
human-judged baseline --- is the right task.  In particular, in many tasks it
is at least as important to avoid {\em inappropriate\/} senses than to select
exactly the right one.  This would be the case in query expansion for
information retrieval, for example, where indiscriminately adding
inappropriate words to a query can degrade performance
\cite{Voorhees:sigir94}.  The examples presented in Section~\ref{sec:examples}
are encouraging in this regard: in addition to performing well at the task of
assigning a high score to the best sense, it does a good job of assigning low
scores to senses that are clearly inappropriate.

Regardless of the criterion for success, the algorithm does need further
evaluation.  Immediate plans include a larger scale version of the experiment
presented here, involving thesaurus classes, as well as a similarly designed
evaluation of how the algorithm fares when presented with noun groups produced
by distributional clustering.  In addition, I plan to explore alternative
measures of semantic similarity, for example an improved variant on simple
path length that has been proposed by Leacock and Chodorow
\shortcite{leacock1994:ms}.

Ultimately, this algorithm is intended to be part of a suite of techniques
used for disambiguating words in running text with respect to WordNet senses.
I would argue that success at that task will require combining knowledge of
the kind that WordNet provides, primarily about relatedness of meaning, with
knowledge of the kind best provided by corpora, primarily about usage in
context.  The difficulty with the latter kind of knowledge is that, until now,
the widespread success in characterizing lexical behavior in terms of
distributional relationships has applied at the level of words --- indeed,
word forms --- as opposed to senses.  This paper represents a step toward
getting as much leverage as possible out of work within that paradigm, and
then using it to help determine relationships among word senses, which is
really where the action is.

\bibliographystyle{authdate}
\begin{small}
% \bibliography{general,learning,distrib,nlstat,ibm_master,senses,ir}

\end{small}

\end{document}